\documentclass{article}

\usepackage{arxiv}

\usepackage[utf8]{inputenc} 
\usepackage[T1]{fontenc}    
\usepackage{hyperref}       
\usepackage{url}            
\usepackage{booktabs}       
\usepackage{amsfonts}       
\usepackage{nicefrac}       
\usepackage{microtype}      
\usepackage{graphicx}
\usepackage{natbib}
\usepackage{doi}

\title{Evaluating Strategies for Synthesizing Clinical Notes for Medical Multimodal AI}

\author{Niccolo Marini\\
Division of Intramural Research, National Library of Medicine, National Institutes of Health\\
Bethesda, MD, 290894, USA\\
{\tt\small niccolo.marini@nih.gov}\thanks{Corresponding author.}
\and
Zhaohui Liang\\
Division of Intramural Research, National Library of Medicine, National Institutes of Health\\
Bethesda, MD, 290894, USA\\
{\tt\small zhaohui.liang@nih.gov}
\and
Sivaramakrishnan Rajaraman\\
Division of Intramural Research, National Library of Medicine, National Institutes of Health\\
Bethesda, MD, 290894, USA\\
{\tt\small sivaramakrishnan.rajaraman@nih.gov}
\and
Zhiyun Xue\\
Division of Intramural Research, National Library of Medicine, National Institutes of Health\\
Bethesda, MD, 290894, USA\\
{\tt\small zhiyun.xue@nih.gov}
\and
Sameer Antani\\
Division of Intramural Research, National Library of Medicine, National Institutes of Health\\
Bethesda, MD, 290894, USA\\
{\tt\small sameer.antani@nih.gov}\footnotemark[1]
}

\date{}

\renewcommand{\shorttitle}{\textit{Evaluating Strategies for Synthesizing Clinical Notes for Medical Multimodal AI}}

\hypersetup{
pdftitle={Evaluating Strategies for Synthesizing Clinical Notes for Medical Multimodal AI},
pdfsubject={Multimodal learning},
pdfauthor={Marini et al.},
pdfkeywords={Multimodal learning, Skin lesion, Co-learning, Synthesized clinical notes, Cross-modality alignment},
}

\begin{document}
\maketitle

\begin{abstract}
Multimodal (MM) learning is emerging as a promising paradigm in biomedical artificial intelligence (AI) applications, integrating complementary modality, which highlight different aspects of patient health.
The scarcity of large heterogeneous biomedical MM data has restrained the development of robust models for medical AI applications. 
In the dermatology domain, for instance, skin lesion datasets typically include only images linked to minimal metadata describing the condition, thereby limiting the benefits of MM data integration for reliable and generalizable predictions.
Recent advances in Large Language Models (LLMs) enable the synthesis of textual description of image findings, potentially allowing the combination of image and text representations. 
However, LLMs are not specifically trained for use in the medical domain, and their naive inclusion has raised concerns about the risk of hallucinations in clinically relevant contexts. 
This work investigates strategies for generating synthetic textual clinical notes, in terms of prompt design and medical metadata inclusion, and evaluates their impact on MM architectures toward enhancing performance in classification and cross-modal retrieval tasks.
Experiments across several heterogeneous dermatology datasets demonstrate that synthetic clinical notes not only enhance classification performance, particularly under domain shift, but also unlock cross-modal retrieval capabilities, a downstream task that is not explicitly optimized during training.
\end{abstract}

\keywords{Multimodal learning \and Skin lesion \and Co-learning \and Synthesized clinical notes \and Cross-modality alignment}

\section{Introduction}
\label{sec:intro}

The growing adoption of multimodal (MM) learning in biomedical deep learning (DL) reflects its potential to integrate heterogeneous sources of information toward improve clinical decision support tools \citep{AFR2022, marini2022unleashing}. The integration of multiple biomedical modalities is critical for capturing complementary aspects of a patient condition \citep{AFR2022}, enabling the early identification of potentially dangerous diseases. However, its full potential is limited by the scarcity of large, well-paired MM datasets \citep{warner2024multimodal}.

Medical images typically encode low-level visual features of disease manifestations, whereas textual clinical notes provide high-level semantic summaries of diagnostic findings \citep{lyu2023multimodal}.
MM learning algorithms can link such complementary perspectives of the patient's health \citep{sun2023scoping}, such as medical imaging and textual clinical notes, for the training of DL models and LLMs toward enriching the quality of the data representation, boosting task performance, and aiding algorithm reliability and generalization across diverse clinical tasks \citep{lyu2023multimodal}, especially in domains where data annotations are time-consuming or inherently heterogeneous \citep{sun2023scoping}.

Despite these promised benefits, the application of MM algorithms remains limited in most biomedical domains \citep{huang2024multimodal, marini2024multimodal}, due to the critical lack of large-scale datasets containing paired samples across the modalities \citep{zhang2023biomedclip}.
Crucially, MM algorithms require paired samples corresponding to the same clinical case, so that models can learn meaningful cross-modal relationships among the features across the modalities \citep{alsaad2024multimodal}.
 
Moreover, MM architectures often require specialized encoders for different data types (e.g., image pixels versus textual tokens), resulting in more complex networks that are prone to overfitting when trained on small datasets \citep{AFR2022}. 
This challenge undermines the generalizability of MM models to unseen data \citep{moor2023foundation}.
While the availability of biomedical data is increasing, many publicly accessible datasets remain unimodal \citep{zhang2023biomedclip}, are annotated with heterogeneous labels \citep{sun2023scoping}, or are insufficiently large, hindering the development and deployment of MM algorithms in clinical settings.
Dermatology domain exemplifies these challenges.
Disease characterization heavily relies on skin lesion clinical and dermoscopic images\citep{weller2014clinical}, with conditions ranging from benign to malignant, including skin cancer, which is the most common cancer in the United States \citep{guy2015vital, guy2015prevalence}.
Despite multiple available dermatology image datasets \citep{zhou2024skincap, yan2025derm1m}, only a few provide paired structured metadata or clinical notes.
Existing MM studies on skin lesion data (\cite{yan2025derm1m}, \cite{zhou2024pre}, \cite{zhou2024skincap}) propose a MM approach integrating dermatology images and clinical notes, relying on large private datasets, which limits advances in the field.
Recent advances in large language models (LLMs) offer a possible solution to these limitations \citep{li2023synthetic}: LLMs can be adopted to generated descriptive clinical notes from minimal information (such as class labels or metadata), enabling the creation of synthetic MM datasets from unimodal image datasets. 
These synthetic clinical notes approximate the information content of real clinical descriptions and can be used to train a MM architecture combining visual and textual information.

However, LLMs trained primarily on general-domain corpora may hallucinate, producing clinically irrelevant or inaccurate descriptions that introduce noise and reduce MM model performance and generalizability \citep{marini2025combining}.
Therefore, the design of synthetic data generation strongly influences the quality and structure of generated text. 
Yet, the impact of such synthetic textual content on MM biomedical model training remains largely unexplored.

This paper investigates how synthesized textual clinical notes influence MM learning when paired with skin lesion images. 
Multiple text generation strategies are evaluated, varying prompt design and metadata usage to guide clinical note synthesis.
The MM architecture performance is assessed on image classification and cross-modal retrieval tasks.

The main contributions of this work are summarized as follows:
\begin{itemize}
    \item A pipeline leveraging image-level metadata and prompt-conditioned LLM generation to produce synthesized clinical notes, enabling MM training when only images are available.
    \item The evaluation of clinical note generation strategies, to analyze the impact of synthesized image descriptions on a MM architecture.
    \item Evaluation of synthetic text effects on eleven heterogeneous test partitions, considering several potential downstream tasks.
\end{itemize}

\section{Methods}
\label{sec:methods}

\subsection{Multimodal Architecture}

\begin{figure}[h]
\centering
\includegraphics[width=0.6\textwidth]{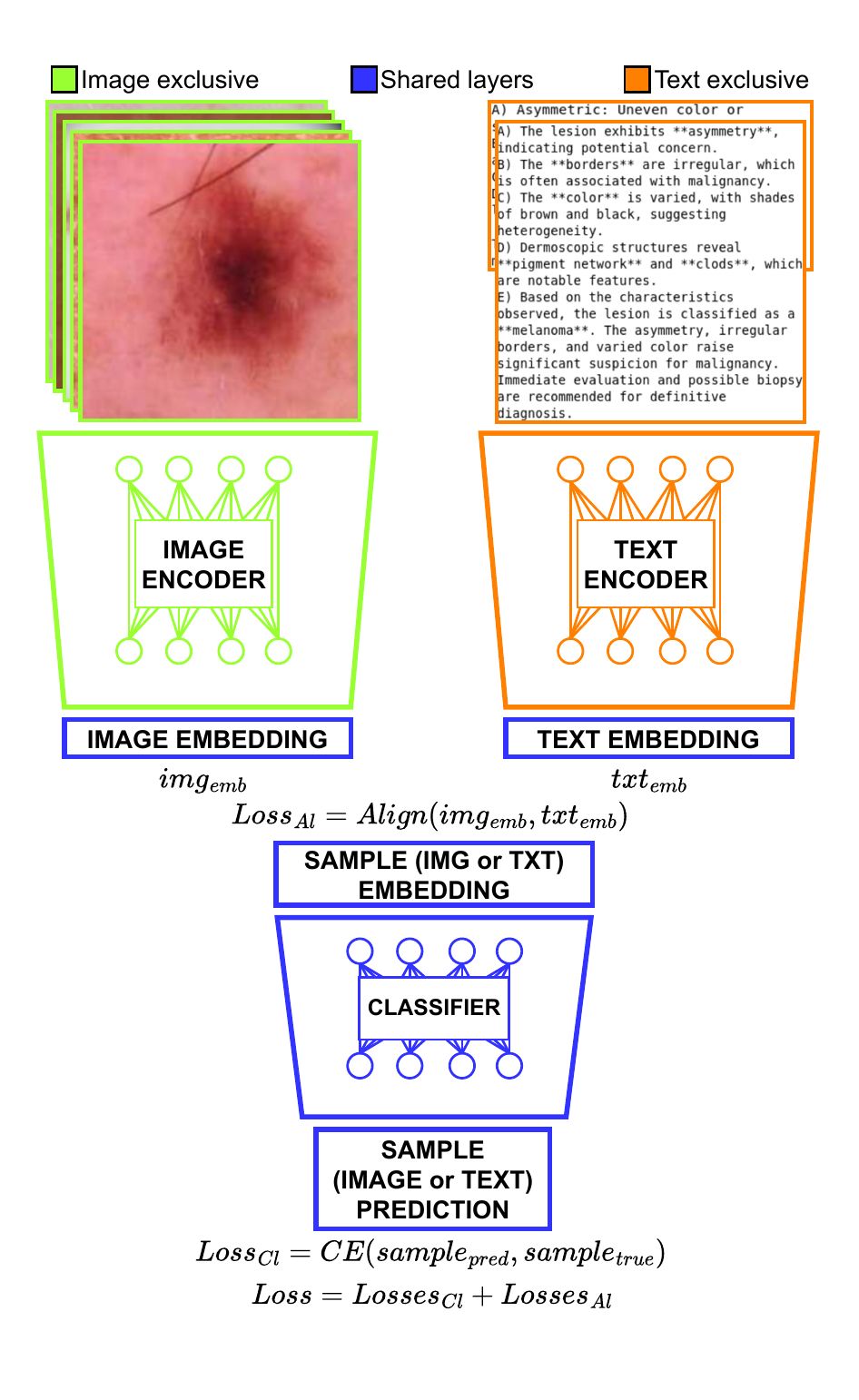} 
\caption{An overview of the proposed MM framework. Skin lesion images and associated textual clinical notes are processed by modality‑specific encoders and projected into a shared representation space that feeds a common classifier. The model is trained to optimize a cross‑entropy term for each modality, together with multiple loss functions (NT\_Xent, L1-Loss and Cosine similarity) to achieve modality alignment, enhancing consistent image–text embeddings. At inference time, either modality can be used on independently.}
\label{fig:architecture}
\end{figure}

The MM architecture is developed to jointly analyze visual and textual dermatology data, exploiting a specific design and a dedicated strategy for cross-modal integration.
Figure \ref{fig:architecture} shows an overview of the model \citep{marini2025combining}.
It includes two input branches with an encoder dedicated to each modality. This is followed by a projection into a shared embedding space and a unified classifier for downstream prediction. 
Each modality is analyzed by a modality specific encoder producing fixed-length embeddings: the image encoder is a Convolutional Neural Network (i.e., DenseNet121, pre-trained with SimCLR \citep{chen2020simple}); the text encoder is a BERT architecture (i.e., PubMedBERT \citep{gu2021domain} which is a transformer model trained on PubMed corpora).
Each encoder produces fixed-length embeddings that are projected into a common latent space, before feeding into a common classifier that processes both types of embeddings.
The classifier is optimized with two cross-entropy terms, images and clinical notes, so that the single representation can be both discriminative on the classification of images.
The strategy for training, as well as classifying input samples, aims to align modalities in the feature space, sharing weights among modalities and adopting specific loss functions during the training: the L1-Loss, the cosine similarity loss and a Self-Supervised loss (i.e. NT-Xent \citep{chen2020simple}). 
The adoption of three loss functions to enhance the alignment acts also as a regularizer that limits overfitting.
The overfitting may represent a problem for the architecture due to several factors: the number of parameters and architectural components that are needed to analyze two different types of data, the optimization of several loss functions in the same training phase, and the limited number of samples available to train the architecture.
The architecture alignment allows the model to perform downstream tasks that are not directly learned during the training phase, such as cross-modal data retrieval.

\subsection{Datasets}
The data used to develop and evaluate the proposed architecture include paired dermatology images and associated textual clinical notes that are collected from multiple public sources.
The data are split into training, validation, and test partitions.
Table \ref{tbl:dataset} summarizes the data distribution across partitions.
During training, the model receives paired image-text samples to learn a shared representation. 
At test time, depending on the task, either a single modality or both modalities may be used. 
This design reflects the intended clinical use case, where more flexibility is required and different tasks may require different setups.
Considering, for example, the classification of images or clinical notes: in clinical practice, experts analyze images and generate clinical notes, so improving the automatic classification of clinical notes offers limited clinical value. 
To enrich robustness and evaluate generalization under distribution shift, data were collected from multiple medical sources, which differ in acquisition protocol, device characteristics, and patient populations. 
Dermatologic images are inherently heterogeneous since lesion appearance varies with magnification, illumination, skin tone, and anatomical site, therefore the adoption of multiple datasets for training provides broader coverage of these factors. 
The test partition includes sources not used for training to assess the out-of-domain performance.
Dermatologic images mostly include two types of images: (i) dermoscopic images, that are high-resolution, magnified views of skin lesions obtained with polarized or non-polarized dermoscopy devices \citep{celebi2019dermoscopy}; and (ii) clinical photographs, that are wider-field, non-magnified images in which the lesion may be not be in the foreground. 
Both image types are common in dermatologic workflows and present similar visual features.
Textual clinical notes used for MM training are generated with multiple strategies, as described in Section \ref{subsec:report_gen}.

\begin{table}[]
\small 
\centering
\caption{Overview of the dataset composition. Data are collected from multiple sources to guarantee heterogeneity. The dataset includes skin lesions and the synthesized clinical notes and it is split into training, validation and testing partitions. The dataset includes five classes: Benign Keratosis (BEK), Benign Nevus (NEV), Actinic Keratosis (ACK), Basal-Cell Cancer (BCC), Melanoma (MEL).}
\resizebox{0.6\textwidth}{!}{
\begin{tabular}{|ccccccc|}
\hline
\multicolumn{1}{|c|}{\textbf{Dataset}} & \multicolumn{1}{c|}{\textbf{BEK}} & \multicolumn{1}{c|}{\textbf{NEV}} & \multicolumn{1}{c|}{\textbf{ACK}} & \multicolumn{1}{c|}{\textbf{BCC}} & \multicolumn{1}{c|}{\textbf{MEL}} & \textbf{Tot} \\ \hline
\multicolumn{7}{|c|}{\textbf{Training partition}}                                                                                                                                                                                         \\ \hline
\multicolumn{1}{|c|}{\textbf{BCN20000}}         & \multicolumn{1}{c|}{796}          & \multicolumn{1}{c|}{2944}         & \multicolumn{1}{c|}{515}          & \multicolumn{1}{c|}{1966}         & \multicolumn{1}{c|}{1999}         & 8220         \\ \hline
\multicolumn{1}{|c|}{\textbf{Derm12345}}        & \multicolumn{1}{c|}{430}          & \multicolumn{1}{c|}{1683}         & \multicolumn{1}{c|}{16}           & \multicolumn{1}{c|}{291}          & \multicolumn{1}{c|}{263}          & 2683         \\ \hline
\multicolumn{1}{|c|}{\textbf{Derm7pt}}          & \multicolumn{1}{c|}{46}           & \multicolumn{1}{c|}{18}           & \multicolumn{1}{c|}{0}            & \multicolumn{1}{c|}{26}           & \multicolumn{1}{c|}{160}          & 250          \\ \hline
\multicolumn{1}{|c|}{\textbf{DermNet}}          & \multicolumn{1}{c|}{171}          & \multicolumn{1}{c|}{39}           & \multicolumn{1}{c|}{38}           & \multicolumn{1}{c|}{133}          & \multicolumn{1}{c|}{62}           & 443          \\ \hline
\multicolumn{1}{|c|}{\textbf{Total}}            & \multicolumn{1}{c|}{1443}         & \multicolumn{1}{c|}{4684}         & \multicolumn{1}{c|}{569}          & \multicolumn{1}{c|}{2416}         & \multicolumn{1}{c|}{2484}         & 11596        \\ \hline
\multicolumn{7}{|c|}{\textbf{Validation partition}}                                                                                                                                                                                       \\ \hline
\multicolumn{1}{|c|}{\textbf{BCN20000}}         & \multicolumn{1}{c|}{170}          & \multicolumn{1}{c|}{630}          & \multicolumn{1}{c|}{110}          & \multicolumn{1}{c|}{421}          & \multicolumn{1}{c|}{428}          & 1759         \\ \hline
\multicolumn{1}{|c|}{\textbf{Derm12345}}        & \multicolumn{1}{c|}{110}          & \multicolumn{1}{c|}{455}          & \multicolumn{1}{c|}{13}           & \multicolumn{1}{c|}{47}           & \multicolumn{1}{c|}{55}           & 680          \\ \hline
\multicolumn{1}{|c|}{\textbf{Derm7pt}}          & \multicolumn{1}{c|}{21}           & \multicolumn{1}{c|}{8}            & \multicolumn{1}{c|}{0}            & \multicolumn{1}{c|}{12}           & \multicolumn{1}{c|}{60}           & 110          \\ \hline
\multicolumn{1}{|c|}{\textbf{DermNet}}          & \multicolumn{1}{c|}{25}           & \multicolumn{1}{c|}{7}            & \multicolumn{1}{c|}{6}            & \multicolumn{1}{c|}{19}           & \multicolumn{1}{c|}{10}           & 67           \\ \hline
\multicolumn{1}{|c|}{\textbf{Total}}            & \multicolumn{1}{c|}{326}          & \multicolumn{1}{c|}{1100}         & \multicolumn{1}{c|}{129}          & \multicolumn{1}{c|}{499}          & \multicolumn{1}{c|}{562}          & 2616         \\ \hline
\multicolumn{7}{|c|}{\textbf{Testing partition}}                                                                                                                                                                                          \\ \hline
\multicolumn{1}{|c|}{\textbf{BCN20000}}         & \multicolumn{1}{c|}{172}          & \multicolumn{1}{c|}{632}          & \multicolumn{1}{c|}{112}          & \multicolumn{1}{c|}{422}          & \multicolumn{1}{c|}{430}          & 1768         \\ \hline
\multicolumn{1}{|c|}{\textbf{Derm12345}}        & \multicolumn{1}{c|}{135}          & \multicolumn{1}{c|}{534}          & \multicolumn{1}{c|}{8}            & \multicolumn{1}{c|}{85}           & \multicolumn{1}{c|}{82}           & 844          \\ \hline
\multicolumn{1}{|c|}{\textbf{Derm7pt}}          & \multicolumn{1}{c|}{13}           & \multicolumn{1}{c|}{5}            & \multicolumn{1}{c|}{0}            & \multicolumn{1}{c|}{7}            & \multicolumn{1}{c|}{45}           & 70           \\ \hline
\multicolumn{1}{|c|}{\textbf{DermNet}}          & \multicolumn{1}{c|}{49}           & \multicolumn{1}{c|}{11}           & \multicolumn{1}{c|}{11}           & \multicolumn{1}{c|}{38}           & \multicolumn{1}{c|}{18}           & 127          \\ \hline
\multicolumn{1}{|c|}{\textbf{HAM10000}}         & \multicolumn{1}{c|}{1338}         & \multicolumn{1}{c|}{7737}         & \multicolumn{1}{c|}{378}          & \multicolumn{1}{c|}{622}          & \multicolumn{1}{c|}{1305}         & 11380        \\ \hline
\multicolumn{1}{|c|}{\textbf{SKINL2}}           & \multicolumn{1}{c|}{33}           & \multicolumn{1}{c|}{97}           & \multicolumn{1}{c|}{0}            & \multicolumn{1}{c|}{40}           & \multicolumn{1}{c|}{28}           & 198          \\ \hline
\multicolumn{1}{|c|}{\textbf{Fitzpatrick17k}}   & \multicolumn{1}{c|}{30}           & \multicolumn{1}{c|}{39}           & \multicolumn{1}{c|}{5}            & \multicolumn{1}{c|}{36}           & \multicolumn{1}{c|}{115}          & 225          \\ \hline
\multicolumn{1}{|c|}{\textbf{Buenos Aires}}     & \multicolumn{1}{c|}{88}           & \multicolumn{1}{c|}{602}          & \multicolumn{1}{c|}{63}           & \multicolumn{1}{c|}{340}          & \multicolumn{1}{c|}{253}          & 1346         \\ \hline
\multicolumn{1}{|c|}{\textbf{SD198}}            & \multicolumn{1}{c|}{60}           & \multicolumn{1}{c|}{32}           & \multicolumn{1}{c|}{62}           & \multicolumn{1}{c|}{13}           & \multicolumn{1}{c|}{38}           & 205          \\ \hline
\multicolumn{1}{|c|}{\textbf{PAD UFES 20}}      & \multicolumn{1}{c|}{52}           & \multicolumn{1}{c|}{68}           & \multicolumn{1}{c|}{13}           & \multicolumn{1}{c|}{64}           & \multicolumn{1}{c|}{26}           & 223          \\ \hline
\multicolumn{1}{|c|}{\textbf{Total}}            & \multicolumn{1}{c|}{1970}         & \multicolumn{1}{c|}{9757}         & \multicolumn{1}{c|}{651}          & \multicolumn{1}{c|}{1667}         & \multicolumn{1}{c|}{2340}         & 16386        \\ \hline
\end{tabular}
}
\label{tbl:dataset}
\end{table}

\subsection{Clinical Notes Generation}
\label{subsec:report_gen}
The clinical notes are synthesized using four strategies that rely on metadata and LLMs. 
Metadata provide information about the lesion class used for classification, such as melanocytic nevus or melanoma, and, in a subset of cases, include subclass annotations that map to one of the five target diagnostic classes (e.g., melanocytic nevus as a subclass of Benign Nevus, or Bowen’s disease as a subclass of Actinic Keratosis).
The first strategy (M) uses only metadata. 
Automatically structured clinical notes are generated by filling predefined templates, such as:
“The image includes a $benign$/$malignant$ skin lesion, specifically a $class$ (specifically a $subclass$),”
where placeholders (benign/malignant, class, subclass) are populated from metadata (the subclass is included only if available).
The other three strategies rely on a Large Language Model (LLM, specifically gpt-4o-mini), where the automatically structured clinical notes act as prompts. 
These prompts request the description of lesion characteristics, including surface structure, color, dermoscopic patterns, and symmetry, in order to enrich the textual content. To minimize hallucinations, each attribute is presented as a set of predefined options, as general-purpose LLMs are not specialized for dermatological interpretation.
The three LLM-based strategies differ in structure, reliance on metadata, and the level of guidance provided:
\begin{itemize}
    \item Prompt 1 (M + P1) focuses on a controlled, template-driven output, requiring the model to select predefined options for key attributes such as lesion symmetry, border type, color, and dermoscopic structures (e.g., pigment network, dots, globules). This approach ensures consistency and limits hallucinations by tightly constraining the vocabulary.
    \item Prompt 2 (M + P2) allows more descriptive freedom while still grounding the text in lesion characteristics—type of lesion (e.g., macule, nodule), color, border definition, and symmetry—without enforcing strict templates.
    \item Prompt 3 (P3) removes all metadata conditioning and prompt the model to independently describe the lesion attributes (proposing a set of attributes) and classify it (proposing a set of possible classes), introducing greater variability and creativity but also a higher risk of hallucinations or inaccurate outputs. The prompt is the same presented in Prompt 1.
\end{itemize}
For clarity, we refer to each multimodal training setup by the name of the synthetic note generation strategy it employs, even if the training of the multimodal architecture does not change depending on the clinical notes used.

\subsection{Data pre-processing}
Data pre-processing involves both images and clinical notes.
The image pre-processing pipeline aims to extract the portion of the image including the lesion and to resize it to 224x224 pixels, as required by CNN input constraints.
The extraction of the portion including the skin lesion depends on the type of image: in the case of dermoscopy images, a mask is generated using Otsu thresholding to identify the lesion region, cropping the portion surrounding it.
In the case of clinical images, a bounding box is manually generated, since it is hard to identify a general rule, due to the heterogeneous acquisition setups of the images.
Textual clinical notes are tokenized into smaller units based on the BERT vocabulary, which contains approximately 30,000 tokens. 
BERT architecture can process input sequences of up to 512 tokens and uses the final hidden state associated with a special token to represent the whole sequence when applied to classification tasks.

\subsection{Experimental Setup}

The experimental setup covers hyperparameter optimization, data encoding, modality-alignment loss functions, data augmentation, and the strategy for addressing class imbalance in the training partition.
Hyperparameters were tuned using a grid search, focusing on the following aspects: number of epochs (15), learning rate (1e-4), and loss function weights (1.0 for all losses, except the self-supervised losses, which is weighted 0.5 NT-Xent). 
The temperature parameter of the SSL losses is set to 0.5 for NT-Xent.
Data encoding varies according to the modality-specific branch. 
For images, a convolutional backbone (DenseNet121) is employed. 
This choice is motivated by the limited size of the dataset and the absence of robust pre-trained Vision Transformers (ViT), which makes CNNs (especially with SimCLR pretraining) more reliable in avoiding overfitting and achieving high-quality image representations.
Loss functions for cross-modal alignment include three components: L1-loss, cosine similarity loss, and the SSL NT\_Xent. 
The L1 and cosine losses enforce sample-wise alignment of feature embeddings within each batch, while the SSL loss encourages better global structure and class-level discrimination across modalities.
Data augmentation is applied to images including rotations, horizontal and vertical flips, and mild RGB perturbations. 
To address class imbalance, underrepresented classes are oversampled, and augmented variants of their samples are preferentially fed into the model.

\section{Results}

\begin{table}[h]
\centering
\caption{Overview of the MM architecture performance on skin lesion classification, considering internal and external test partitions. Results are evaluated on $\kappa$-score, comparing unimodal (Img) and multimodal models. MM models correspond to the same architecture, trained with different synthesized clinical notes: M (only metadata), M + P1 (metadata and LLM prompt 1), M + P2 (metadata and LLM prompt 2), M + P3 (LLM prompt 3).}
\resizebox{0.9\textwidth}{!}{

\begin{tabular}{|cccccc|}
\hline
\multicolumn{1}{|c|}{\textbf{Dataset}}        & \multicolumn{1}{c|}{\textbf{Img}}      & \multicolumn{1}{c|}{\textbf{M}}                 & \multicolumn{1}{c|}{\textbf{M + P1}}            & \multicolumn{1}{c|}{\textbf{M + P2}}            & \textbf{P3}                \\ \hline
\multicolumn{6}{|c|}{\textbf{Internal partition}}                                                                                                                                                                                                                         \\ \hline
\multicolumn{1}{|c|}{\textbf{BCN20000}}       & \multicolumn{1}{c|}{$0.686 \pm 0.018$} & \multicolumn{1}{c|}{$0.712 \pm 0.025$}          & \multicolumn{1}{c|}{$0.709 \pm 0.018$}          & \multicolumn{1}{c|}{$0.711 \pm 0.017$}          & \textbf{$0.717 \pm 0.011$} \\ \hline
\multicolumn{1}{|c|}{\textbf{Derm12345}}      & \multicolumn{1}{c|}{$0.719 \pm 0.016$} & \multicolumn{1}{c|}{$0.733 \pm 0.018$}          & \multicolumn{1}{c|}{$0.735 \pm 0.025$}          & \multicolumn{1}{c|}{$0.729 \pm 0.022$}          & \textbf{$0.736 \pm 0.020$} \\ \hline
\multicolumn{1}{|c|}{\textbf{Derm7pt}}        & \multicolumn{1}{c|}{$0.586 \pm 0.068$} & \multicolumn{1}{c|}{$0.622 \pm 0.052$}          & \multicolumn{1}{c|}{\textbf{$0.624 \pm 0.051$}} & \multicolumn{1}{c|}{$0.567 \pm 0.045$}          & $0.588 \pm 0.084$          \\ \hline
\multicolumn{1}{|c|}{\textbf{DermNet}}        & \multicolumn{1}{c|}{$0.725 \pm 0.026$} & \multicolumn{1}{c|}{$0.735 \pm 0.044$}          & \multicolumn{1}{c|}{$0.712 \pm 0.036$}          & \multicolumn{1}{c|}{$0.710 \pm 0.034$}          & \textbf{$0.736 \pm 0.032$} \\ \hline
\multicolumn{6}{|c|}{\textbf{External partition}}                                                                                                                                                                                                                         \\ \hline
\multicolumn{1}{|c|}{\textbf{HAM10000}}       & \multicolumn{1}{c|}{$0.470 \pm 0.014$} & \multicolumn{1}{c|}{$0.479 \pm 0.025$}          & \multicolumn{1}{c|}{$0.480 \pm 0.023$}          & \multicolumn{1}{c|}{\textbf{$0.482 \pm 0.012$}} & $0.468 \pm 0.029$          \\ \hline
\multicolumn{1}{|c|}{\textbf{SKINL2}}         & \multicolumn{1}{c|}{$0.686 \pm 0.094$} & \multicolumn{1}{c|}{$0.738 \pm 0.050$}          & \multicolumn{1}{c|}{$0.730 \pm 0.049$}          & \multicolumn{1}{c|}{\textbf{$0.743 \pm 0.050$}} & $0.706 \pm 0.056$          \\ \hline
\multicolumn{1}{|c|}{\textbf{Fitzpatrick17k}} & \multicolumn{1}{c|}{$0.418 \pm 0.053$} & \multicolumn{1}{c|}{$0.449 \pm 0.046$}          & \multicolumn{1}{c|}{$0.436 \pm 0.043$}          & \multicolumn{1}{c|}{$0.444 \pm 0.045$}          & \textbf{$0.456 \pm 0.038$} \\ \hline
\multicolumn{1}{|c|}{\textbf{Buenos Aires}}   & \multicolumn{1}{c|}{$0.402 \pm 0.064$} & \multicolumn{1}{c|}{$0.413 \pm 0.065$}          & \multicolumn{1}{c|}{$0.401 \pm 0.075$}          & \multicolumn{1}{c|}{\textbf{$0.441 \pm 0.062$}} & $0.417 \pm 0.059$          \\ \hline
\multicolumn{1}{|c|}{\textbf{SD198}}          & \multicolumn{1}{c|}{$0.518 \pm 0.039$} & \multicolumn{1}{c|}{$0.501 \pm 0.037$}          & \multicolumn{1}{c|}{$0.486 \pm 0.030$}          & \multicolumn{1}{c|}{$0.501 \pm 0.036$}          & \textbf{$0.547 \pm 0.049$} \\ \hline
\multicolumn{1}{|c|}{\textbf{PAD UFES 20}}    & \multicolumn{1}{c|}{$0.335 \pm 0.053$} & \multicolumn{1}{c|}{\textbf{$0.384 \pm 0.064$}} & \multicolumn{1}{c|}{$0.378 \pm 0.051$}          & \multicolumn{1}{c|}{$0.369 \pm 0.040$}          & $0.367 \pm 0.048$          \\ \hline

\end{tabular}
}
\label{tbl:classification}
\end{table}

The MM architecture outperforms the unimodal baselines in the classification of dermatology images for both internal and external testing partitions, and achieves robust performance on cross-modal retrieval when trained with clinical notes generated using metadata, which is a task not directly optimized during model training.

This section presents a quantitative evaluation of the proposed MM architecture, which integrates real skin lesion images with synthesized clinical notes generated through different prompting strategies: M (metadata only), M + P1 (metadata with LLM prompt 1), M + P2 (metadata with LLM prompt 2), and M + P3 (LLM prompt 3 without metadata). 
Results are grouped by internal vs. external test partitions to distinguish the performance on data collected from the same sources used for training and data collected from different, inherently more heterogeneous sources.
For clarity, we refer to each multimodal training setup by the name of the synthetic note generation strategy it employs, even if the training of the multimodal architecture does not change depending on the clinical notes used.
Two tasks are considered: (i) image classification, where unimodal image-based training (Img) serves as the baseline, and (ii) cross-modal retrieval, which evaluates the alignment of image and text embeddings under different strategies, despite the retrieval task was not explicitly optimized during training.
Therefore, the architecture is trained with the same approach, changing only the textual descriptions.

The combination of images and textual clinical notes improves the average performance on the image classification, since every MM strategy matches or exceeds the corresponding unimodal baseline across all datasets.
Table \ref{tbl:classification} summarizes the image classification performance in terms of Cohen’s $\kappa$-score, considering both internal and external datasets. 
A higher $\kappa$-score indicates stronger agreement with ground-truth labels beyond chance.
Considering the internal partition (in-distribution evaluation), the MM architecture trained with strategy M improves or matches image-only across all four internal datasets.  
The clinical notes generated with LLMs produce mixed but competitive effects: P3 leads to high performance, such as M + P1 and M + P2, even if their performance is more prone to variability. 
Larger internal datasets (BCN20000, Derm12345) exhibit relatively low variability, so modest absolute gains are likely meaningful. 
On the other hand, smaller or more heterogeneous sets (Derm7pt, DermNet) show higher standard deviations.
Considering the external partition (out-of-distribution evaluation), the performance generally drops in comparison with the internal partitions.
Yet, the MM architecture leads to higher performance. 
All six external datasets show at least one MM strategy outperforming the architecture trained only with images.
SKINL2 is the dataset that benefits most from the MM training approach.
Clinical notes synthesized with M + P2 generally allow to build the most robust model for the classification of external partition data.
In general, the strategy exploiting only metadata (the most simple) allows to reach good performance in several datasets. 
LLM prompts add incremental but inconsistent value, reaching good performance, sometimes the highest, but also showing sometimes lower performance.

\begin{table}[h]
\centering
\caption{Overview of the MM architecture performance on skin lesion clinical note retrieval, using images as input. The clinical notes are retrieved from a pool of samples including all four kind of textual descriptions. Results are evaluated on mAP. MM models correspond to the same architecture, trained with different synthesized clinical notes.}

\resizebox{0.8\textwidth}{!}{
\begin{tabular}{|ccccc|}
\hline
\multicolumn{1}{|c|}{\textbf{Dataset}}        & \multicolumn{1}{c|}{\textbf{M}}        & \multicolumn{1}{c|}{\textbf{M + P1}}   & \multicolumn{1}{c|}{\textbf{M + P2}}            & \textbf{P3}       \\ \hline
\multicolumn{5}{|c|}{\textbf{Internal partition}}                                                                                                                                                     \\ \hline
\multicolumn{1}{|c|}{\textbf{BCN20000}}       & \multicolumn{1}{c|}{$0.724 \pm 0.006$} & \multicolumn{1}{c|}{\textbf{$0.787 \pm 0.012$}} & \multicolumn{1}{c|}{$0.782 \pm 0.009$} & $0.510 \pm 0.006$ \\ \hline
\multicolumn{1}{|c|}{\textbf{Derm12345}}      & \multicolumn{1}{c|}{$0.795 \pm 0.009$} & \multicolumn{1}{c|}{\textbf{$0.827 \pm 0.015$}} & \multicolumn{1}{c|}{$0.821 \pm 0.005$} & $0.646 \pm 0.012$ \\ \hline
\multicolumn{1}{|c|}{\textbf{Derm7pt}}        & \multicolumn{1}{c|}{\textbf{$0.771 \pm 0.004$}} & \multicolumn{1}{c|}{$0.723 \pm 0.018$} & \multicolumn{1}{c|}{$0.721 \pm 0.018$} & $0.465 \pm 0.013$ \\ \hline
\multicolumn{1}{|c|}{\textbf{DermNet}}        & \multicolumn{1}{c|}{$0.735 \pm 0.007$} & \multicolumn{1}{c|}{\textbf{$0.748 \pm 0.021$}} & \multicolumn{1}{c|}{$0.742 \pm 0.012$} & $0.404 \pm 0.009$ \\ \hline
\multicolumn{5}{|c|}{\textbf{External partition}}                                                                                                                                                     \\ \hline
\multicolumn{1}{|c|}{\textbf{HAM10000}}       & \multicolumn{1}{c|}{$0.631 \pm 0.009$} & \multicolumn{1}{c|}{\textbf{$0.761 \pm 0.014$}} & \multicolumn{1}{c|}{$0.750 \pm 0.006$} & $0.620 \pm 0.010$ \\ \hline
\multicolumn{1}{|c|}{\textbf{SKINL2}}         & \multicolumn{1}{c|}{$0.692 \pm 0.010$} & \multicolumn{1}{c|}{$0.777 \pm 0.016$} & \multicolumn{1}{c|}{\textbf{$0.787 \pm 0.015$}} & $0.588 \pm 0.008$ \\ \hline
\multicolumn{1}{|c|}{\textbf{Fitzpatrick17k}} & \multicolumn{1}{c|}{$0.564 \pm 0.005$} & \multicolumn{1}{c|}{\textbf{$0.570 \pm 0.014$}} & \multicolumn{1}{c|}{$0.564 \pm 0.017$} & $0.413 \pm 0.010$ \\ \hline
\multicolumn{1}{|c|}{\textbf{Buenos Aires}}   & \multicolumn{1}{c|}{$0.537 \pm 0.009$} & \multicolumn{1}{c|}{\textbf{$0.601 \pm 0.015$}} & \multicolumn{1}{c|}{$0.590 \pm 0.023$} & $0.479 \pm 0.007$ \\ \hline
\multicolumn{1}{|c|}{\textbf{SD198}}          & \multicolumn{1}{c|}{$0.534 \pm 0.008$} & \multicolumn{1}{c|}{$0.571 \pm 0.012$} & \multicolumn{1}{c|}{\textbf{$0.577 \pm 0.013$}} & $0.416 \pm 0.004$ \\ \hline
\multicolumn{1}{|c|}{\textbf{PAD UFES 20}}    & \multicolumn{1}{c|}{$0.392 \pm 0.007$} & \multicolumn{1}{c|}{\textbf{$0.421 \pm 0.015$}} & \multicolumn{1}{c|}{$0.418 \pm 0.012$} & $0.274 \pm 0.003$ \\ \hline
\end{tabular}
}
\label{tbl:ret_img}
\end{table}

Tables \ref{tbl:ret_img} and \ref{tbl:ret_txt} present the cross-modal retrieval results, reported in terms of mean Average Precision (mAP), where one modality is used as a query to retrieve samples from the other.
Since the retrieval task is not explicitly optimized during training, it provides a robust evaluation of the flexibility of the MM architecture across diverse and heterogeneous clinical tasks.
We evaluate two retrieval setups: (i) clinical note retrieval (\ref{tbl:ret_img}) and (ii) image retrieval (\ref{tbl:ret_txt}).

Table \ref{tbl:ret_img} presents the results on the cross-modal retrieval task considering skin lesion images used as input to retrieve synthesized clinical notes. 
The pool of samples to retrieve includes clinical notes synthesized with all the possible strategies.
The findings show that combining metadata with LLM-based prompts (M + P1 and M + P2) consistently improves performance compared to using metadata alone (M). 
The benefit is especially clear considering the external partitions, where the performance gap is more noticeable.
Although the performance is generally lower when relying solely on metadata-generated notes (M), this strategy still outperforms approaches using notes synthesized without metadata (P3). 
This finding highlights a key result: models trained with the P3 clinical notes reach lower results on the cross-modal retrieval.
This suggests that notes generated without metadata tend to be noisier and more prone to hallucinations, which weakens cross-modal alignment and limits their usefulness in downstream multimodal applications. 
In contrast, structured prompts incorporating metadata provide richer clinical notes, enabling stronger alignment between modalities.
The result contrasts with the trends observed in image classification tasks, where P3 leads to some of the highest performance in terms of classification.

\begin{table}[h]
\centering
\caption{Overview of the MM architecture performance on skin lesion image retrieval, using the clinical notes as input. Each column corresponds to a different text generation strategy used during training: M (metadata only), M + P1 (metadata with LLM prompt 1), M + P2 (metadata with LLM prompt 2), and P3 (LLM prompt without metadata). Furthermore, each column reports results evaluated on clinical notes that were generated using the same strategy as the one used for training that model (e.g., the M column is evaluated on notes generated with strategy M, the column M + P1 on the clinical notes generated with M + P1, etc.). Performance is reported in terms of mean Average Precision (mAP). All MM models share the same architecture but differ in the synthesized clinical notes used for training}
\resizebox{0.8\textwidth}{!}{
\begin{tabular}{|ccccc|}
\hline
\multicolumn{1}{|c|}{\textbf{Dataset}}        & \multicolumn{1}{c|}{\textbf{M}}        & \multicolumn{1}{c|}{\textbf{M + P1}}   & \multicolumn{1}{c|}{\textbf{M + P2}}            & \textbf{P3}       \\ \hline
\multicolumn{5}{|c|}{\textbf{Internal partition}}                                                                                                                                                     \\ \hline
\multicolumn{1}{|c|}{\textbf{BCN20000}}       & \multicolumn{1}{c|}{$0.996 \pm 0.004$} & \multicolumn{1}{c|}{$0.976 \pm 0.009$} & \multicolumn{1}{c|}{\textbf{$0.953 \pm 0.004$}} & $0.393 \pm 0.004$ \\ \hline
\multicolumn{1}{|c|}{\textbf{Derm12345}}      & \multicolumn{1}{c|}{$0.990 \pm 0.004$} & \multicolumn{1}{c|}{$0.987 \pm 0.008$} & \multicolumn{1}{c|}{\textbf{$0.964 \pm 0.004$}} & $0.503 \pm 0.012$ \\ \hline
\multicolumn{1}{|c|}{\textbf{Derm7pt}}        & \multicolumn{1}{c|}{$0.989 \pm 0.008$} & \multicolumn{1}{c|}{$0.973 \pm 0.014$} & \multicolumn{1}{c|}{\textbf{$0.969 \pm 0.007$}} & $0.431 \pm 0.023$ \\ \hline
\multicolumn{1}{|c|}{\textbf{DermNet}}        & \multicolumn{1}{c|}{$0.974 \pm 0.007$} & \multicolumn{1}{c|}{$0.954 \pm 0.017$} & \multicolumn{1}{c|}{\textbf{$0.942 \pm 0.010$}} & $0.310 \pm 0.010$ \\ \hline
\multicolumn{5}{|c|}{\textbf{External partition}}                                                                                                                                                     \\ \hline
\multicolumn{1}{|c|}{\textbf{HAM10000}}       & \multicolumn{1}{c|}{$0.792 \pm 0.028$} & \multicolumn{1}{c|}{$0.761 \pm 0.024$} & \multicolumn{1}{c|}{\textbf{$0.933 \pm 0.003$}} & $0.469 \pm 0.015$ \\ \hline
\multicolumn{1}{|c|}{\textbf{SKINL2}}         & \multicolumn{1}{c|}{$0.955 \pm 0.003$} & \multicolumn{1}{c|}{$0.959 \pm 0.010$} & \multicolumn{1}{c|}{\textbf{$0.955 \pm 0.004$}} & $0.480 \pm 0.007$ \\ \hline
\multicolumn{1}{|c|}{\textbf{Fitzpatrick17k}} & \multicolumn{1}{c|}{$0.970 \pm 0.004$} & \multicolumn{1}{c|}{$0.977 \pm 0.012$} & \multicolumn{1}{c|}{\textbf{$0.927 \pm 0.005$}} & $0.422 \pm 0.015$ \\ \hline
\multicolumn{1}{|c|}{\textbf{Buenos Aires}}   & \multicolumn{1}{c|}{$0.995 \pm 0.004$} & \multicolumn{1}{c|}{$0.836 \pm 0.021$} & \multicolumn{1}{c|}{\textbf{$0.958 \pm 0.003$}} & $0.438 \pm 0.007$ \\ \hline
\multicolumn{1}{|c|}{\textbf{SD198}}          & \multicolumn{1}{c|}{$0.909 \pm 0.013$} & \multicolumn{1}{c|}{$0.865 \pm 0.021$} & \multicolumn{1}{c|}{\textbf{$0.939 \pm 0.004$}} & $0.372 \pm 0.006$ \\ \hline
\multicolumn{1}{|c|}{\textbf{PAD UFES 20}}    & \multicolumn{1}{c|}{$0.839 \pm 0.017$} & \multicolumn{1}{c|}{$0.921 \pm 0.013$} & \multicolumn{1}{c|}{\textbf{$0.903 \pm 0.009$}} & $0.308 \pm 0.005$ \\ \hline
\end{tabular}
}
\label{tbl:ret_txt}
\end{table}

Table \ref{tbl:ret_txt} presents the results on the cross-modal retrieval task considering clinical notes used as input to retrieve skin lesion images. 
Each column corresponds to a specific generation strategy, and the input notes are produced using the same strategy on which the model is evaluated. 
For instance, the first column (M) evaluates the performance of the model trained with clinical notes generated with the M strategy, using the same type of clinical notes as input to retrieve skin lesion images.
This setup assumes that the architecture trained on notes generated with a given strategy achieves optimal performance when evaluated on inputs generated by the same strategy. 
The results demonstrate the effectiveness of the MM architecture for cross-modal retrieval of images. 
Considering the strategies using notes synthesized incorporating metadata (M), the models achieve near-perfect performance, with mAP values exceeding 0.90 in several datasets, including when tested on external partitions. 
In particular, performance on external datasets remains generally strong, highlighting the robustness of structured note synthesis. 
This indicates that the synthesized notes are highly informative and well-aligned with the corresponding images. 
However, the finding does not apply to the P3 strategy, which generates notes without metadata.
The MM architecture trained with those notes consistently reaches lower mAP values (0.308–0.480), reflecting the presence of noisier, less semantically aligned content. 
This result confirms what was experienced on the cross-modal retrieval of images: clinical notes synthesized without metadata are noisy and lead to poor alignment across modalities.

The fact that modalities are poorly aligned is confirmed when evaluating the similarity among paired samples.
Table \ref{tbl:dist} presents the evaluation on the alignment across modalities, showing the cosine-similarity between image feature embeddings and the feature embeddings from the corresponding textual description.
The Table follows a similar structure to the previous one (Table \ref{tbl:ret_txt}): each column corresponds to a specific generation strategy and it is evaluated on that specific type of synthesized notes.
As hypothesized, the model can align images with the corresponding textual description generated with the same strategy.
For each strategy adopted to synthesize clinical notes with metadata (i.e. M, M + P1, M + P2), paired image-note couples are well aligned (cosine similarity close to 1). 
However, also in this case, the poorest performance is reached on the architecture trained on P3.
This finding confirms that image encoders trained with clinical notes generated without any metadata are poorly aligned across modalities, suggesting that the lack of structured metadata in P3 leads to descriptions that struggle to capture medical details and hallucinate.

\begin{table}[h]
\centering
\caption{Overview of the alignment between image and (synthesized) textual clinical notes. For each column, the textual descriptions are synthesized using the strategy corresponding to the column header. Results are evaluated on cosine similarity. MM models correspond to the same architecture, trained with different synthesized clinical notes.}
\resizebox{0.8\textwidth}{!}{
\begin{tabular}{|ccccc|}
\hline
\multicolumn{1}{|c|}{\textbf{Dataset}}        & \multicolumn{1}{c|}{\textbf{M}}        & \multicolumn{1}{c|}{\textbf{M + P1}}   & \multicolumn{1}{c|}{\textbf{M + P2}}            & \textbf{P3}       \\ \hline
\multicolumn{5}{|c|}{\textbf{Internal partition}}                                                                                                                                                     \\ \hline
\multicolumn{1}{|c|}{\textbf{BCN20000}}       & \multicolumn{1}{c|}{$0.735 \pm 0.014$} & \multicolumn{1}{c|}{$0.717 \pm 0.028$} & \multicolumn{1}{c|}{\textbf{$0.689 \pm 0.018$}} & $0.230 \pm 0.012$ \\ \hline
\multicolumn{1}{|c|}{\textbf{Derm12345}}      & \multicolumn{1}{c|}{$0.797 \pm 0.008$} & \multicolumn{1}{c|}{$0.775 \pm 0.014$} & \multicolumn{1}{c|}{\textbf{$0.747 \pm 0.014$}} & $0.258 \pm 0.021$ \\ \hline
\multicolumn{1}{|c|}{\textbf{Derm7pt}}        & \multicolumn{1}{c|}{$0.696 \pm 0.023$} & \multicolumn{1}{c|}{$0.684 \pm 0.036$} & \multicolumn{1}{c|}{\textbf{$0.667 \pm 0.021$}} & $0.312 \pm 0.035$ \\ \hline
\multicolumn{1}{|c|}{\textbf{DermNet}}        & \multicolumn{1}{c|}{$0.722 \pm 0.011$} & \multicolumn{1}{c|}{$0.705 \pm 0.036$} & \multicolumn{1}{c|}{\textbf{$0.679 \pm 0.032$}} & $0.275 \pm 0.019$ \\ \hline
\multicolumn{5}{|c|}{\textbf{External partition}}                                                                                                                                                     \\ \hline
\multicolumn{1}{|c|}{\textbf{HAM10000}}       & \multicolumn{1}{c|}{$0.479 \pm 0.046$} & \multicolumn{1}{c|}{$0.452 \pm 0.036$} & \multicolumn{1}{c|}{\textbf{$0.640 \pm 0.011$}} & $0.230 \pm 0.019$ \\ \hline
\multicolumn{1}{|c|}{\textbf{SKINL2}}         & \multicolumn{1}{c|}{$0.683 \pm 0.026$} & \multicolumn{1}{c|}{$0.691 \pm 0.031$} & \multicolumn{1}{c|}{\textbf{$0.688 \pm 0.030$}} & $0.234 \pm 0.018$ \\ \hline
\multicolumn{1}{|c|}{\textbf{Fitzpatrick17k}} & \multicolumn{1}{c|}{$0.472 \pm 0.032$} & \multicolumn{1}{c|}{$0.472 \pm 0.046$} & \multicolumn{1}{c|}{\textbf{$0.458 \pm 0.037$}} & $0.308 \pm 0.026$ \\ \hline
\multicolumn{1}{|c|}{\textbf{Buenos Aires}}   & \multicolumn{1}{c|}{$0.451 \pm 0.043$} & \multicolumn{1}{c|}{$0.380 \pm 0.043$}  & \multicolumn{1}{c|}{\textbf{$0.448 \pm 0.042$}} & $0.204 \pm 0.025$ \\ \hline
\multicolumn{1}{|c|}{\textbf{SD198}}          & \multicolumn{1}{c|}{$0.428 \pm 0.026$} & \multicolumn{1}{c|}{$0.441 \pm 0.033$} & \multicolumn{1}{c|}{\textbf{$0.465 \pm 0.036$}} & $0.223 \pm 0.016$ \\ \hline
\multicolumn{1}{|c|}{\textbf{PAD UFES 20}}    & \multicolumn{1}{c|}{$0.291 \pm 0.043$} & \multicolumn{1}{c|}{$0.313 \pm 0.045$} & \multicolumn{1}{c|}{\textbf{$0.286 \pm 0.048$}} & $0.207 \pm 0.026$ \\ \hline
\end{tabular}
}
\label{tbl:dist}
\end{table}


\section{Discussion \& Conclusion}

The proposed MM algorithm enhances the representation of dermatologic data by combining information from real images with synthesized clinical notes, despite the possibility of hallucinations in the latter.

Images are inherently high-dimensional and unstructured, where individual pixels require contextual information for semantic meaning. 
In contrast, short and pertinent clinical notes provide compact and interpretable semantic representations. 
Aligning these modalities enables knowledge transfer from text to images, improving the understanding of visual content and strengthening interpretability and generalizability. 
This is confirmed by the improvement in classification performance of the MM architecture compared to unimodal baselines. 
Beyond classification, modality alignment also enables new downstream tasks with emergent capabilities, such as cross-modal retrieval, even though these were not explicitly optimized during training.

When clinical notes are synthesized from metadata and/or LLM prompts, there remains a risk of hallucination that introduces noisy content and hinders the learning process. 
This work analyzes the effect of different prompts on MM training. 
Counterintuitively, prompts P3 (no metadata) perform well in classification tasks despite lacking metadata for their synthesis, possibly because label supervision dominates during training. 
However, P3 underperforms in retrieval tasks, suggesting that free-form LLM-generated notes introduce hallucinations or irrelevant details that weaken the alignment between text and images. 
These prompts leave many degrees of freedom to the LLM: lesions are not linked to any class, forcing the model to generate labels or sublabels, while morphological details also lack structured options, encouraging hallucination. 
Conversely, metadata-guided prompts M + P1 and especially M + P2 achieve stronger retrieval performance across datasets, highlighting the importance of structured, clinically grounded text for robust MM embeddings.

The MM representation is richer and more robust than the unimodal one, achieving higher performance on both internal and external datasets, although results on external partitions remain more variable. 
Stronger performance of M + P2 across internal and external partitions indicates that combining metadata with structured prompts improves resilience to distribution shifts. 
The analysis of alignment further confirms this trend: models trained with metadata-driven prompts, particularly M + P2, not only align well with their own note structure but also generalize effectively across other prompt styles, reflecting greater semantic robustness. 
Importantly, the source of the note—real or synthetic—matters less than the inclusion of contextually relevant content. 
This is reinforced by the underperformance of P3, even on its own note type, showing that hallucinated or weakly grounded content introduces noise that disrupts image-text correspondence.

Overall, MM training acts as a form of implicit regularization, enriching the image encoder with semantic knowledge that supports both generalization and tasks beyond classification. 
This is particularly relevant in clinical practice, where retrieval tasks could assist case-based reasoning and decision support, with a possible impact on clinical practice. 
For example, medical students may retrieve cases based on text or images for learning, while experts could compare new cases with prior ones to obtain textual summaries of relevant information.

Despite these advances, limitations remain. 
The synthesis of clinical notes carries a persistent risk of hallucinations, which are not reviewed by experts and may propagate noise or errors into training. 
Even limited hallucinations can weaken cross-modal alignment, undermining retrieval performance and practical clinical utility. 
This limitation arises in part because the LLM is not trained on medical corpora, which could reduce noise. 
The architecture itself also has constraints: it relies on standard CNN and BERT branches, which are not novel. 
This choice, however, reflects the limited dataset size and the need to avoid overfitting.

Future work should explore domain-specific prompt engineering, hybrid real-synthetic note generation, and adaptive retrieval strategies to better exploit the complementary strengths of image and text.
Further directions include combining prompts to reduce LLM-induced noise, evaluating notes generated by medically trained LLMs, and developing new architectures that can be trained with fewer annotations.

\bibliographystyle{unsrtnat}
\bibliography{references}  

\end{document}